\documentclass[journal]{IEEEtran}
\usepackage{amsmath,graphicx}
\usepackage{algorithm}
\usepackage{algorithmic}
\usepackage{amsmath,amssymb,amsfonts}
\usepackage{framed,multirow}
\usepackage{xcolor}

\usepackage[para,online,flushleft]{threeparttable}
\begin{document}
\title{Medical Instrument Segmentation in 3D US by Hybrid Constrained Semi-Supervised Learning}
\author{Hongxu Yang$^*$, Caifeng Shan$^*$, Arthur Bouwman, \\ Lukas R. C. Dekker, Alexander F. Kolen and Peter H. N. de With
\thanks{Hongxu Yang (h.yang@tue.nl, corresponding author) and Peter H. N. de With are with the Department
of Electrical Engineering, Eindhoven University of Technology, Eindhoven, The Netherlands.}
\thanks{Caifeng Shan (caifeng.shan@gmail.com, corresponding author) is with Shandong University of Science and Technology, Qingdao, China.}
\thanks{Arthur Bouwman and Lukas R.C. Dekker are with Catharina Hospital, Eindhoven, The Netherlands.}
\thanks{Alexander F. Kolen is with Philips Research, Eindhoven, The Netherlands.}}

\maketitle

\begin{abstract}
Medical instrument segmentation in 3D ultrasound is essential for image-guided intervention. However, to train a successful deep neural network for instrument segmentation, a large number of labeled images are required, which is expensive and time-consuming to obtain. In this article, we propose a semi-supervised learning (SSL) framework for instrument segmentation in 3D US, which requires much less annotation effort than the existing methods. To achieve the SSL learning, a Dual-UNet is proposed to segment the instrument. The Dual-UNet leverages unlabeled data using a novel hybrid loss function, consisting of uncertainty and contextual constraints. Specifically, the uncertainty constraints leverage the uncertainty estimation of the predictions of the UNet, and therefore improve the unlabeled information for SSL training. In addition, contextual constraints exploit the contextual information of the training images, which are used as the complementary information for voxel-wise uncertainty estimation. Extensive experiments on multiple \emph{ex-vivo} and \emph{in-vivo} datasets show that our proposed method {achieves Dice score of about 68.6\%-69.1\% and the inference time of about 1 sec. per volume. These results are better than the state-of-the-art SSL methods and the inference time is comparable to the supervised approaches.}
\end{abstract}


\IEEEpeerreviewmaketitle
\section{Introduction}
Advanced imaging modalities, such as X-ray and ultrasound (US), have been widely applied during the minimally invasive intervention and surgery. Cardiac interventions, such as RF-ablation therapy and cardiac TAVI procedures, require manipulation of the instrument and US probe inside the patient body to reach the target area and perform the intended operation. 3D US imaging is an attractive modality to guide instrument during cardiac interventions because of its real-time radiation-free image visualization capacity and rich spatial information for tissue anatomy and instruments. Furthermore, the US imaging system is low-cost and provides mobility at the  hospital. However, 3D US imaging faces challenges of lower image resolution, and low image contrast between tissue and artifacts, which require experienced sonographers to interpret the US data during intervention. Sometimes, efforts is spent to detect the instrument rather than performing the operation itself due to the complicated multi-coordinate alignment. Therefore, automatic instrument segmentation in 3D US images is greatly desired in the computer-assisted intervention.

{Medical instrument segmentation is to build a mapping function from the input US to a 3D segmentation mask, which is mostly solved by deep neural networks because of their powerful feature representation \cite{DLReview}. In practice, medical instrument segmentation suffers the challenges of limited training images with a tiny portion of the image being instrument voxels (instrument typically occupies 0.01-0.1\% voxels in the volume), which commonly leads to less accurate segmentation results when applying a one-step segmentation strategy \cite{man2019deep}. Consequently, a coarse-to-fine segmentation method \cite{MEDIA2020} is proposed. In the first stage, the instrument region is roughly localized based on the coarse segmentation results, then the fine segmentation is performed in the instrument region. Therefore, the final segmentation performance heavily relies on the output of the first stage. Nevertheless, because of the limited training images, the scarce annotation, and low contrast US imaging, it is challenging to train the detection network for instrument localization. In addition, even with a correctly coarse localization of the instrument, it is challenging to train a fine segmentation network given limited annotated 3D US images, where it would take several hours for an experienced expert to annotated one US volume.}
\subsection{Related Work}
In this section, we first introduce the related work of instrument segmentation in 3D US images. Then, previous works about semi-supervised learning methods are discussed.

\subsubsection{Instrument segmentation in 3D US}
Instrument segmentation in the 3D US is usually treated as a voxel-wise classification task, assigning a semantic category to every 3D voxel or region. Machine learning methods with handcrafted features were proposed to train the supervised learning classifier, which exhaustively classifies voxels for instrument segmentation by sliding window \cite{pourtaherian2017improving}. Nevertheless, these approaches rely on experience to design the handcrafted features, which limits the capacity of discriminating information for accurate instrument segmentation. Recently, Deep learning (DL) methods \cite{DLReview} have been proposed to segment the instrument in 3D US by voxel-wise classification \cite{IJCARS2019}. However, these methods have limitations like compromised semantic information usage by decomposing 3D patches into tri-planar slices and computationally-expensive predictions. To address the limitations of the tri-planar strategy, a 3D patch-based method was introduced to exploit the 3D information~\cite{MEDIA2020}. To avoid exhaustive voxel classification, the coarse-to-fine strategy was considered to improve the segmentation efficiency~\cite{chen2019med3d,li2018h,MEDIA2020}. Another challenge for medical instrument segmentation in the 3D US is the size of the 3D US images for the limited GPU memory, which hampers the network with satisfactory inference efficiency. Arif \emph{et al.} \cite{arif2019automatic} proposed an efficient UNet~\cite{ronneberger2015u} for instrument segmentation, but their simplified network cannot handle the US images with complex anatomical structures \cite{MEDIA2020}. Based on the Yang \emph{et al.}~\cite{MEDIA2020}, an efficient and accurate segmentation solution is a \emph{coarse-to-fine strategy} for medical instrument segmentation~\cite{chen2019med3d}, i.e., first detecting the interested area and then performing fine segmentation on the interested area. Nevertheless, these methods apply a fully supervised learning in both coarse and fine stages, which require annotated data that are expensive and laborious to obtain. Therefore, these methods are not feasible on a large scale dataset with annotation challenges.

\subsubsection{Semi-supervised learning} 
Semi-supervised learning (SSL) methods~\cite{MICCAI2017Zhang,MICCAI2018Nie,BMVC2018,MASSL,MICCAI2019SSL1,MICCAI2019SSL2} have been studied for medical image segmentation, which reduce the annotation efforts for CNN training and leverage abundant unlabeled images. The most popular SSL methods follow consistency enforcing strategy \cite{MT, liu2020semi}, which leverage the unlabeled data  by constraining the network predictions to be consistent under perturbations in input or network parameters. A typical example is the student-teacher model, a specific application of knowledge distillation strategy \cite{wang2020knowledge}. Specifically, the teacher-student model was proposed to distillate the prediction distribution knowledge from a complex model (so-called teacher), which is then used to train a simplified and faster model (commonly denoted as student) \cite{hinton2015distilling}. The recent SSL methods exploit the teacher-student approach rather than the above knowledge distillation~\cite{gong2018teaching}, which train a teacher model based on labeled images, and then the labeled and unlabeled images' predictions from the teacher model are used as supervision for the student model training. However, for standard teacher-student model, teacher model cannot learn unlabeled images' information, which may lead to unstable predictions for student supervision. Alternatively, the mean-teacher (MT) \cite{MT} model exploits the unlabeled information in both teacher and student models simultaneously, which achieves state-of-the-art performance in a variety of applications. Nevertheless, several limitations exist for a standard MT model in segmentation tasks. First, a typical MT model expects to minimize the distance between the predictions from two models \cite{MT}. However, direct distance measure without prediction selection would lead to network degradation, which can be confused by too many less confident sample points. As a result, it is challenging for image segmentation tasks with lots of unconfident prediction points. Meanwhile, the soft information of predicted results is not adequately exploited because of a simple measurement. Second, the temporal parameters' updating in MT leads to information correlation, which unfortunately introduces the knowledge bias~\cite{DualStudent}. To address the above issues, several solutions were proposed recently. An uncertainty-aware self-ensembling model was proposed~\cite{MICCAI2019SSL1,MICCAI2019SSL2} to make use of certainty estimations for the segmentation of unlabeled images, which enhances the segmentation performance with limited annotations. Although uncertainty-aware methods \cite{MICCAI2019SSL1,MICCAI2019SSL2} achieve superior performances, they are all based on the mean-teacher approach with exponential moving averaging (EMA) on parameter updating, which still encounters a parameter-correlation problem between teacher and student models. To overcome the network weight bias from EMA, Dual-Student was proposed to perform interactive prediction refinement between two parallel student models~\cite{DualStudent}. Although the Dual-Student achieves a better performance than an MT method, it only exploits discriminative information for image classification, which may not be sufficient for semantic segmentation. 
\subsection{Our Work}
By considering the challenges of annotation in both coarse and fine stages, and inspired by Dual-Student~\cite{DualStudent}, we proposed a deep Q-network (DQN) driven Dual-UNet framework, which was preliminary validated in our initial MICCAI work~\cite{MICCAI2020Yang}. It aims to improve the overall segmentation efficiency with minimized annotation effort, in both coarse localization and fine segmentation stages by employing reinforcement learning and semi-supervised learning. In this work, to better exploit unlabeled contextual information, we further 
improve our earlier work with a more advanced and well-defined contextual constraint w.r.t. label-wise and network-wise design. Specifically, the proposed hybrid constraint exploits voxel-level uncertainty information and contextual-level similarities between the predictions of Dual-UNet, which leverages the discriminating information of unlabeled images. Extensive experiments show that the Dual-UNet achieves state-of-the-art segmentation performance by leveraging a small amount of labeled training images with abundant unlabeled images. In addition to the initial validation on a challenging \emph{ex-vivo} dataset~\cite{MICCAI2020Yang}, two extra \emph{in-vivo} datasets are included in this article. Extensive ablation studies and comparisons with the SOTA methods are presented in this article, with more implementation details and quantitative and qualitative results. 

In summary, this paper presents the following contributions:
\begin{itemize}
\item {An annotation efficient medical instrument segmentation method is proposed based on semi-supervised learning. The proposal is able to exploit unlabeled information (non-voxel annotation), which leverages abundant unlabeled images for instrument segmentation.}
\item {We propose a hybrid constraint for SSL learning, which exploits the unsupervised signal in both voxel and contextual level. Therefore, the unlabeled information can be better exploited for SSL learning.}
\item {The proposed method is thoroughly evaluated on multiple challenging datasets, \emph{ex-vivo} RF-ablation catheter dataset, \emph{in-vivo} TAVI guide-wire dataset, and an \emph{in-vivo} validation dataset. The results show the effectiveness of our method and its potential for clinical applications.}
\end{itemize}

The remainder of this paper is organized as follows. Our method is described in Section~\ref{methods}. The experiments and results are presented in Section~\ref{experiments} and~\ref{results}, respectively. Section \ref{discussions} discusses the limitation of the paper. Finally, the paper is concluded in Section \ref{conclusions}.

\begin{figure}[htbp]
\centering{\includegraphics[width=9cm]{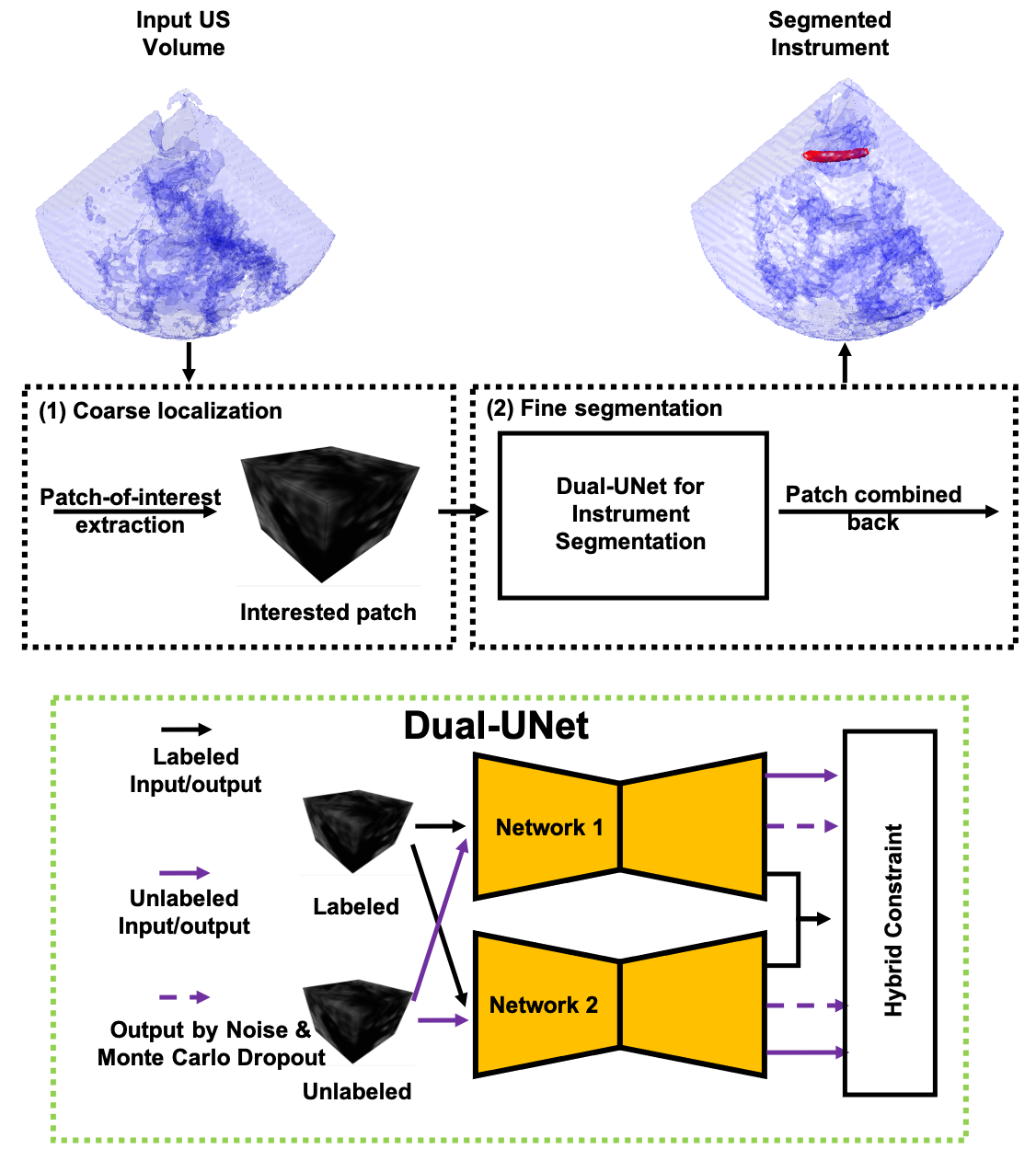}} 
\caption{Schematic view of the proposed framework. (1) The input 3D volumetric data is processed by a coarse localization algorithm, which localizes the instrument center point in 3D space. (2) Local patches around the detected points are extracted and segmented by the Dual-UNet, which is trained by the proposed SSL scheme. The output of Dual-UNet is the average result of two predictions. The prediction patches are combined to generate the final prediction output.}
\label{overview}
\end{figure}

\section{Our Method}\label{methods}
As shown in Fig.~\ref{overview}, the proposed coarse-to-fine instrument segmentation framework includes two stages. First, the instrument's location is obtained by a coarse locator. Second, the Dual-UNet, trained by the SSL framework, is applied on local patches around the estimated location for fine instrument segmentation. Following our previous publication in MICCAI 2020~\cite{MICCAI2020Yang}, DQN is adopted as the pre-selector, which can efficiently localize the interested region of the instrument in 3D US by a policy learned off-line. By doing so, accurate voxel-level annotation can be avoided by only considering the instrument center point in the space. Other coarse localization method can also be considered, such as Single Shot Multibox Detector (SSD) \cite{liu2016ssd} and Faster R-CNN \cite{ren2015faster}. Nevertheless, these methods need more complex annotations of the bounding box and larger training datasets. In the following sections, based on the pre-selection of the instrument region, a fine segmentation method is proposed based on the SSL framework. More details of DQN pre-selection can be founded in our MICCAI paper~\cite{MICCAI2020Yang}.
\subsection{Semi-supervised Dual-UNet for segmentation}
With the coarse localization of the instrument in 3D US, the instrument is then segmented by the proposed patch-based Dual-UNet, which is trained by a hybrid constrained SSL framework. Given the training patches containing $N$ labeled patches $\{(x_i,y_i)\}_{i=1}^{N}$ and $M$ unlabeled patches $\{x_j\}_{j=1}^{M}$, where $x\in{\mathbb{R}}^{V^3}$ is the 3D input patch and $y\in\{0,1\}^{V^3}$ is the corresponding annotation (where $V^3$ is the size of the image or patch), the task is to minimize the following hybrid loss function:
\begin{equation}\label{Lossoverview}
\mathcal{L}_{\text{hybrid}}=L_{\text{sup}}+L_{\text{semi}},
\end{equation} 
{where the $L_\text{sup}$ means the standard supervised loss and $L_\text{semi}$ represents the proposed constraints for semi-supervised learning. They are gradually introduced as follows.}
\subsubsection{Supervised loss function $L_{\text{sup}}$}
In this paper, we consider the standard cross-entropy and Dice hybrid loss function as the supervised loss. Given the label $y$ and its corresponding prediction $\hat{y}$, the $L_{\text{sup}}$ is defined as
\begin{equation}\label{Lsup}
\scriptsize
L_{\text{sup}}=-\sum[y_i\log(\hat{y}_i)+(1-y_i)\log(1-\hat{y}_i)]+[1-\frac{2\sum{y_i}\hat{y_i}+1}{\sum{y_i}+\sum\hat{y}_i+1}],
\end{equation} 
where the first term is binary cross-entropy and the second term is Dice loss~\cite{sudre2017generalised}. Specifically, $i$ is the index of the voxels in the image and $\sum$ means sum of all voxels.
\subsubsection{Semi-supervised loss function $L_{\text{semi}}$}
\begin{figure}[htbp]
\centering{\includegraphics[width=6cm]{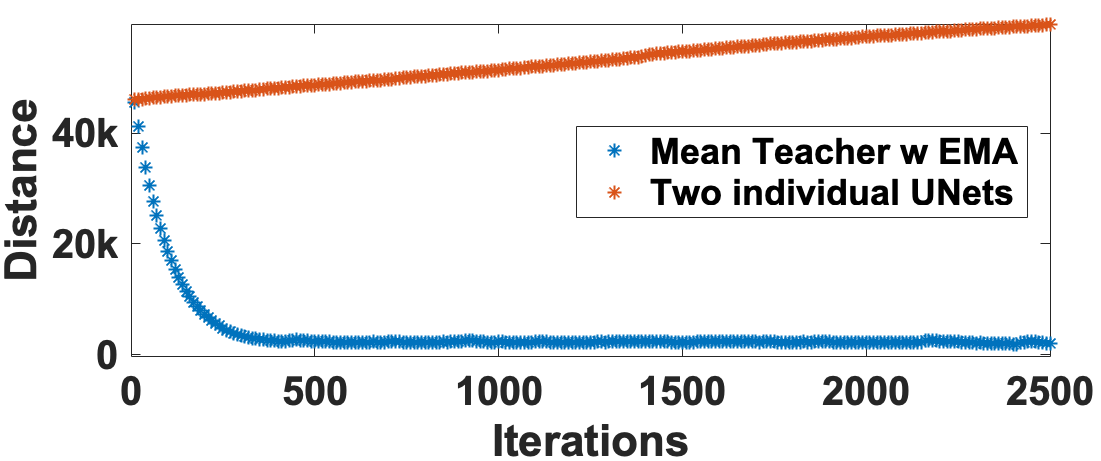}} 
\caption{Sum of parameters' distance of two UNets in Mean-Teacher with EMA and two individual UNets from our scheme. {The distance is measured by the summation of the paired weight distance: $\sum_i|\theta_1^i-\theta_2^i|$, where $i$ is the index of the filter weight and \{1,2\} are two networks in mean-teacher and Dual-UNet.}}
\label{Distance}
\end{figure}

To exploit the unlabeled image under the supervised signal from labeled data, we propose an SSL training scheme based on a novel hybrid constraint, which employs a Dual-UNet as the segmentation network. The proposed Dual-UNet structure is motivated by the Dual-Student framework for classification~\cite{DualStudent}, which is different to mean-teacher methods. Conventionally, the mean-teacher method learns the network parameters by updating a student network from a teacher network \cite{MICCAI2019SSL1,MICCAI2019SSL2}. Intuitively, mean-teacher method introduces two parallel networks whose parameters are highly correlated due to performing an exponential moving average (EMA) on the updating process. As a result, the obtained knowledge is biased and may not be discriminative enough \cite{DualStudent}. Alternatively, our proposed Dual-UNet utilizes two independent networks to learn the discriminating information by knowledge interaction through uncertainty constraints. The parameters distance in Fig.~\ref{Distance} indicates, our scheme does not have the parameter correlation problem of EMA-mean-teacher, thus avoiding the two networks to be the same. In addition, these two networks learn knowledge from each other without any domination. 

Compared to Dual-Student, which only classifies the images with limited information exploitation, our Dual-UNet can segment the images by an advanced hybrid constraint. Specifically, the hybrid constraint consists of two types of elements to exploit the information at different level: voxel-level constraint and contextual-level constraint:
\begin{enumerate}
\item
\emph{Voxel-level constraint:} An intra-network uncertainty constraint ($L_\text{intra}$) and an inter-network uncertainty constraint ($L_\text{inter}$) are defined to exploit the voxel-wise discriminating information of the unlabeled images' prediction. These two constraints are based on the predictions' uncertainty estimation, which select the most confident predictions as the supervised signal. Therefore, the most reliable samples of the unlabeled images are communicated between two individual networks, which forces the networks to generate similar predictions with different parameter values. 
\item
\emph{Context-level constraint:} Label-wise constraint ($L_\text{LCont}$) and Network-wise constraint ($L_\text{NCont}$) are introduced to exploit the semantic information between un-/labeled predictions, and contextual similarities between networks' prediction, respectively. Because these two constraints are exploiting the semantic information within predictions and annotations, they can be complementary information for voxel-level uncertainty estimations. 
\end{enumerate}
Details of the hybrid constraint components are shown below.
\\
\noindent {{\textbf {Intra-network Uncertainty Constraint $L_{\text{intra}}$:}}} {Although there are some literature directly using the prediction from network to guide the unsupervised learning \cite{BMVC2018,wang2020knowledge}, the direct usage of the predictions might include noisy and misclassified voxels, which leads to unsatisfactory results.} To generate reliable predictions from history and {use them to} guide the network to learn discriminating information gradually, we design an uncertainty constraint for each network. Given an input patch, $T$ predictions are generated by T times forward passes, based on a Monte Carlo Dropout (MCD) and input with Gaussian noise (GN) \cite{MC}. Therefore, the estimated probability map for a class is obtained by the average of $T$ times prediction for an input patch, resulting in $\hat{P}$ for each network. Based on the above probability maps, the uncertainty of this map is measured by $\hat{U}=-\sum_{c}\hat{P}log(\hat{P})$ for $c$ different classes and the loss constraint for a network is formulated by:
\begin{equation}\label{LC}
L_\text{intra}=\frac{\sum(\mathcal{I}(\hat{U}<\tau_1)\odot||\hat{y}-\hat{P}||)}{\sum\mathcal{I}(\hat{U}<\tau_1)},
\end{equation}
where $\sum$ is the sum of all the voxels. $\mathcal{I}$ is a binary indicator function, $\tau_1$ is a threshold to measure the uncertainty \cite{MICCAI2019SSL2}, which selects the most reliable voxels by binary voxel-level multiplication $\odot$. Parameter $\hat{y}$ is the prediction for a network. By following this approach, the proposed strategy is approximately equal to the mean-teacher method with the history step as unity in the methods \cite{MICCAI2019SSL1,MICCAI2019SSL2}. Intuitively speaking, this constraint selects the reliable voxels from Bayesian predictions, where only the most confident points are selected to guide the network.
\\
\noindent {{\textbf {Inter-network Uncertainty Constraint $L_{\text{inter}}$:}}} Besides the above uncertainty constraint for each network, we also propose an uncertainty constraint to measure the prediction consistency between two individual networks to constrain the knowledge and avoid bias \cite{DualStudent}. The proposed inter-network uncertainty constraint {lets the networks to learn the discriminating information} by comparing the predictions between two networks with stable voxel selection. With the above definitions of normal prediction ($\hat{y}$) and averaged Bayesian prediction ($\hat{P}$), their corresponding binary predictions are obtained as $C$ and $\hat{C}$, respectively, which are thresholded by 0.5 for fair class distribution. Based on these, more stable voxels (i.e., less uncertain) for each network is defined as
\begin{equation}\label{Stable}
\mathcal{S}=\mathcal{I}(C\odot\hat{C})\odot(\mathcal{I}(U<\tau_2)\oplus\mathcal{I}(\hat{U}<\tau_2)),
\end{equation}
where $U$ is the uncertainty based on normal output and $\hat{U}$ is the uncertainty based on Bayesian output. $\tau_2$ is a stronger threshold to select the more stable voxels for Network $q$ than Eqn.~(\ref{LC}). By using a voxel-based logical OR ($\oplus$), stable instrument voxels are loosely selected to find the matched prediction voxels from the same-class prediction. Furthermore, we also define the voxel-level probability distance $D=||\hat{y}-\hat{P}||$, which indicates the predictions' consistency. With definitions of stable voxels and probability distances, the less stable voxels in the stable samples are optimized to enhance {the overall voxel confidence between two networks}. Specifically, the inter-network uncertainty constraint $L_\text{inter}$ for Network~1 is formulated by:
\begin{equation}\label{LS}
\scriptsize
L_\text{inter}=\frac{\sum(((\mathcal{S}_1\odot\mathcal{S}_2\odot\mathcal{I}(D_1>D_2))\oplus(\overline{\mathcal{S}_1\odot\mathcal{S}_2}\odot\mathcal{S}_2))\odot||\hat{y}_1-\hat{y}_2||)}{\sum((\mathcal{S}_1\odot\mathcal{S}_2\odot\mathcal{I}(D_1>D_2))\oplus(\overline{\mathcal{S}_1\odot\mathcal{S}_2}\odot\mathcal{S}_2))},
\end{equation}
where the subscripts in the $\mathcal{S}$ and $D$ represent the serial number of networks in the Dual-UNet, $||\cdot||$ is the probability distance at the voxel level by norm-2 and $\overline{(\cdot)}$ is a binary NOT operation. Intuitively, the operation $\mathcal{S}_1\odot\mathcal{S}_2\odot\mathcal{I}(D_1>D_2)$ selects the less stable voxels from Network~1 by comparing the probability distance from the two networks' stable voxels. As for function $\overline{\mathcal{S}_1\odot\mathcal{S}_2}\odot\mathcal{S}_2$, if the voxels are not stable for Networks 1 but are stable for Network 2, then these voxels' information are used to guide the Network 1 to generate a similar prediction. This uncertainty constraint enables the unsupervised signal communication between two individual network, and train the Network~1. A similar expression with mirrored indexes applies to Network~2.
\begin{figure}[htbp]
\centering{\includegraphics[width=8cm]{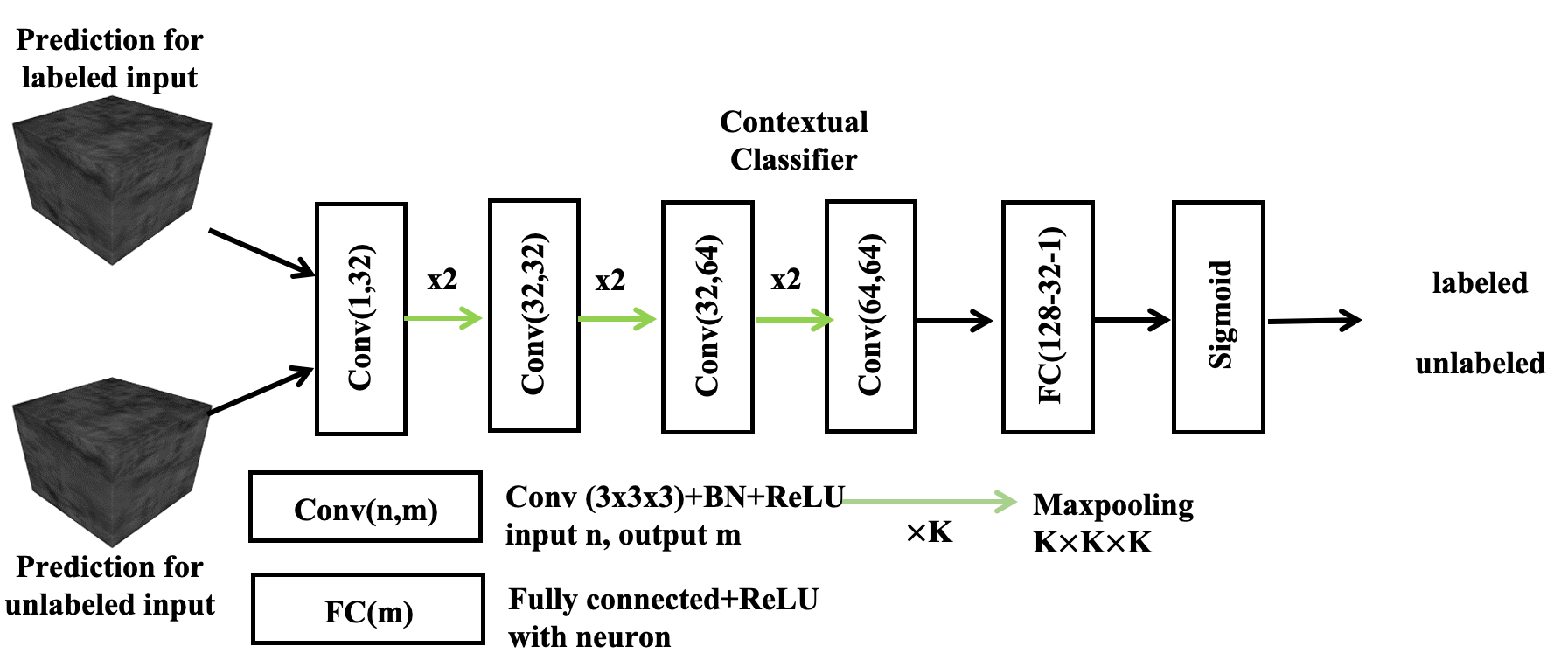}} 
\caption{Architecture of the proposed classifier for $L_{\text{LCont}}$. The network distinguishes the input is labeled or not.}
\label{classifier}
\end{figure}

\noindent {{\textbf {Label-wise Contextual Constraint $L_{\text{LCont}}$:}}} The above loss constraints on the intra-/inter-network consider voxel-level consistency of paired predictions, i.e., the predictions from two networks for the same input while ignoring the differences between labeled and unlabeled predictions at the contextual level (due to unlabeled predictions has no annotation to learn). Intuitively, to learn the prediction consistency at the whole input level, we also introduce a contextual constraint based on the implementation of adversarial learning. Specifically, the labeled and unlabeled predictions are analyzed by a classifier, as shown in Fig.~\ref{classifier}, to generate the image class: labeled or unlabeled, which are used to generate the binary cross-entropy. $L_{\text{LCont}}$ is defined as
\begin{equation}\label{clss}
L_{\text{LCont}}=Cls\log(\hat{Cls})+(1-Cls)\log(1-\hat{Cls}),
\end{equation}
where the $\hat{Cls}$ is the predicted class whether the input prediction has corresponding annotation or not, while $Cls$ is the prior knowledge of the prediction has annotation or not. The negative sign is considered to maximize the similarity between the labeled predictions and unlabeled predictions, while the loss function is minimized to distinguish them.

\noindent {{\textbf {Network-wise Contextual Constraint $L_{\text{NCont}}$:}}} The label-wise contextual constraint focuses on the contextual difference between labeled predictions and unlabeled predictions, which ignores the contextual information consistency between two individual networks. To fully exploit this contextual information at the network-level, a network-wise contextual constraint is introduced. Specifically, it has different processing steps for labeled and unlabeled predictions. (1) The labeled images' predictions and corresponding annotation are processed by an encoder to generate contextual vectors, which are used to measure the latent space similarity between prediction and annotation. (2) As for the unlabeled predictions from the two networks, their contextual vectors are measured to enforce themselves to be as similar as possible. The contextual encoder (CE) has a similar structure with that in Fig.~\ref{classifier}, but excludes the FC layers and adds one extra Conv layer (kernel number of 64), the $L_{\text{NCont}}$ is defined as the vector distance by norm-2:
\begin{equation}\label{context}
\scriptsize
L_{\text{NCont}}=||CE(\hat{y}_1^l)-CE(y)||+||CE(\hat{y}_2^l)-CE(y)||+||CE(\hat{y}_1^u)-CE(\hat{y}_2^u)||,
\end{equation}
where $\hat{y}$ and $y$ are predictions and corresponding annotation. $l$ and $u$ represent labeled and unlabeled patches. This network-wise constraint compensates the intra-network contextual information usage in $L_{\text{LCont}}$ and enforces the information interaction similar to $L_\text{inter}$. Based on the design, the $CE$ is trained properly from the supervised shape signal, which is simultaneously used to enforce the unlabeled predictions from different UNets to be the same.

{Based on the above constraint definition, the SSL loss function for both networks is aggregated as follows:}
\begin{equation}\label{Loss}
\mathcal{L}_{\text{semi}}=\alpha(L_{\text{intra}}+L_{\text{inter}})+\beta{L_{\text{LCont}}}+\gamma{L_{\text{NCont}}},
\end{equation} 
where coefficients $\alpha$, $\beta$ and $\gamma$ are parameters to balance the weight between different components. It is worth to mention that the above hybrid loss function is applied to both networks during the training. {In addition, the derivatives analysis of the above components are shown in Appendix section.}

Intuitively, $L_\text{sup}$ uses labeled information to guide the networks to converge to correct predictions and optimize the direction in hyper-parameter space. In contrast to supervised information, $L_\text{intra}$ focuses on each individual network's information uncertainty. Specifically, it considers MCD and GN to generate noised and less confident predictions, which are processed by uncertainty estimation to select the reliable predictions in the patch. With these selected voxels, the probability distance between normal predictions to these voxels is minimized to enhance the network's confidence, which avoids the voxels with less confidence or noise from a common $\Pi$-model. However, in contrast to the uncertainty-aware network \cite{MICCAI2019SSL1,MICCAI2019SSL2}, which employed two separate networks with historical parameter correlation, our method ensembles two networks into one with history step as unity. Moreover, instead of intra-network information usage, $L_\text{inter}$ focuses on voxel-level uncertainty interactions, which omits the parameter correlations and generates more diverse network parameters from the random initialization, MCD and GN. In detail, it is designed to select more stable voxels based on predictions, which are used to reduce the probability distance between the predictions of these stable voxels from two networks. As described in definitions, $L_{\text{LCont}}$ is considered to maximize the prediction similarity between labeled and unlabeled outputs, which enforces the unlabeled predictions to gradually similar to labeled predictions at contextual level. In contrast $L_{\text{NCont}}$ is used to enforce a higher contextual similarity between the two networks' predictions.
\section{Experiments}\label{experiments}
\subsection{Datasets and Preprocessing}
\begin{figure}[htbp]
\centering{\includegraphics[width=7cm]{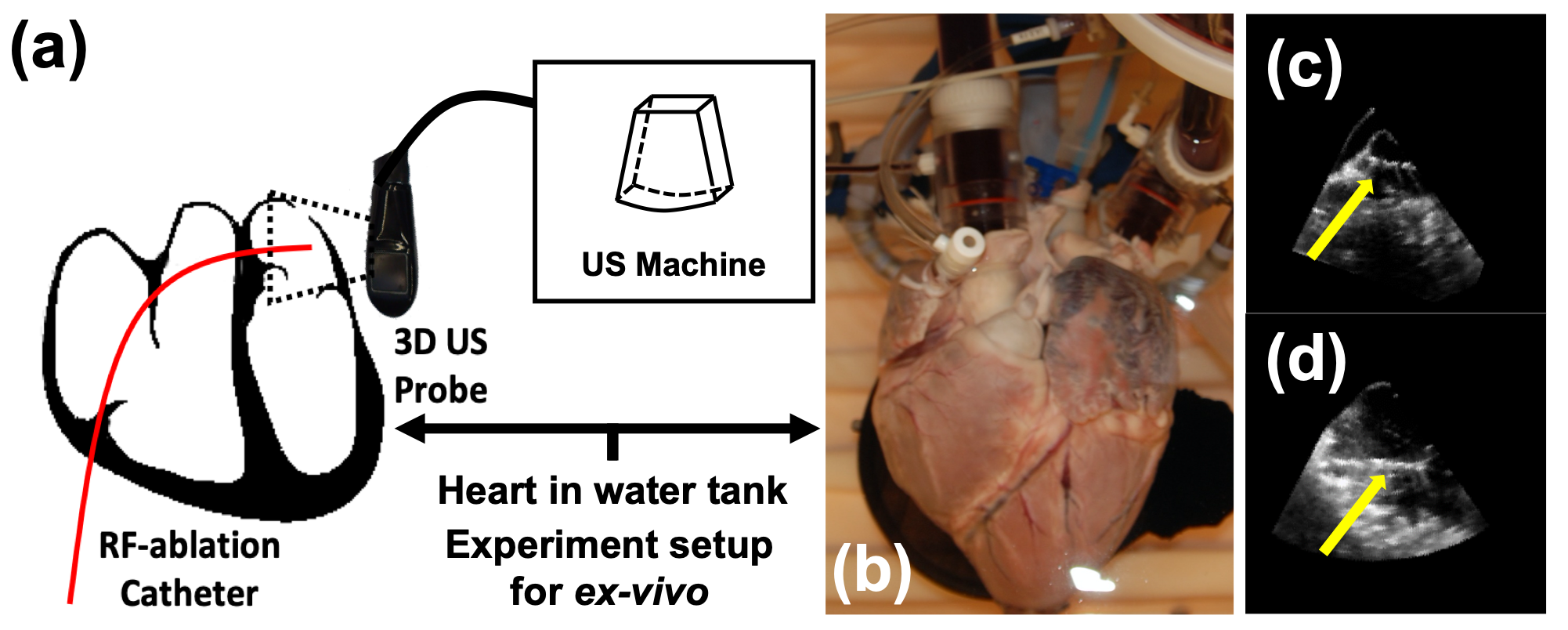}} 
\caption{(a) Recording setup for the \emph{ex-vivo} dataset. (b) A isolated porcine heart was placed in a water tank with the catheter inserted into a chamber. (c)(d) Slices from 3D images, where the yellow arrows point at the catheter.}
\label{dataset}
\end{figure}

\noindent {{\textbf {\emph{Ex-vivo} RF-ablation catheter dataset:}}} To validate our instrument segmentation method, we collected an \emph{ex-vivo} dataset on RF-ablation catheter for cardiac intervention, consisting of 88 3D cardiac US volumetric data from eight porcine hearts. During the recording, the heart was placed in a water tank with the RF-ablation catheter (with the diameter of 2.3-3.3 mm) inside the heart chambers. The phased-array US probe (X7-2t with 2,500 elements by Philips Medical Systems, Best, Netherlands) was placed next to the interested chambers to capture the images containing the catheter, which was monitored by a US console (EPIQ 7 by Philips). For each recording, we pulled out the catheter, and re-inserted it into the heart chamber, and placed the probe with different locations and view angles, to minimize the overlap among images. Therefore, the recorded images avoid information leakage for network training and evaluations. The recording setup and example slices are shown in Fig.~\ref{dataset}. As can be observed, the catheter has similar intensity dynamic range to the tissue, which makes it challenging to segment the catheter. The obtained volumetric images are re-sampled to the volume size of $160\times160\times160$ voxels (where padding is applied at the boundary to make the volume such that it has the equal size in each direction), which leads to a voxel size range of 0.3-0.8 mm. All the volumes were manually annotated at the voxel level. To validate the proposed method, 60 volumes were randomly selected for training, 7 volumes were used as the validation images, and 21 volumes were used as the testing images. To train the Dual-UNet, 6, 12, and 18 of 60 volumes were used as the labeled images, while the remainder were the unlabeled for SSL training. \\

\noindent {{\textbf {\emph{In-vivo} RF-ablation catheter dataset:}}} To further validate the generalization of the proposed method, an \emph{in-vivo} RF-ablation catheter dataset was collected from two live porcine hearts, which includes 13 images with RF-ablation catheter in the heart chambers. The data collection was approved by the ethical committee and recorded at GDL, Utrecht University, the Netherlands (ID: AVD115002015205, 2015-07-03). The images were collected by a phased-array US probe (X7-2t with 2,500 elements by Philips Medical Systems, Best, Netherlands). During the recording, medical doctor manipulated the catheter to reach the different region of the heart chamber, where the RF-ablation procedures were performed. Similar to the \emph{ex-vivo} dataset, the images are re-sampled to the volume size of $160\times160\times160$ voxels. The obtained images were manually annotated at the voxel-level. All these 13 volumes are used to validate the generalization of the model, which was trained on the above \emph{ex-vivo} dataset.
\\
\noindent {{\textbf {\emph{In-vivo} TAVI guide-wire dataset:}}} We also collected an \emph{in-vivo} TAVI guide-wire dataset including eighteen volumes from two TAVI operations. The study was approved by the institutional review board of Philips (ICBE) and the Catharina Hospital Eindhoven (Medical Research Ethics Committees United, MEC-U; study ID:non-WMO 2017-106). Patients approved the use of anonymous data for retrospective analysis. During the recording, the sonographer recorded images from different locations of the chamber without interfering the procedure. The volumes were recorded with a mean volume size of $201\times202\times302$ voxels. Similar to the above \emph{ex-vivo} dataset, volumes are re-sampled to have a volume size of $160\times160\times160$ voxels. The guide-wire (0.889 mm) has a diameter of around 3-5 voxels due to spatial distortion. The images were manually annotated by clinical experts to generate the binary segmentation mask as the annotation. We randomly divided the dataset into three parts: 12 volumes for training, 2 volumes for validation and 4 volumes for testing. Specifically for the training images, 2, 4, and 6 volumes of 12 images were selected as the labeled images, while the rest were used as the unlabeled images for SSL training. 

\subsection{Implementation Details and Training Process}
\begin{figure}[htbp]
\centering{\includegraphics[width=9cm]{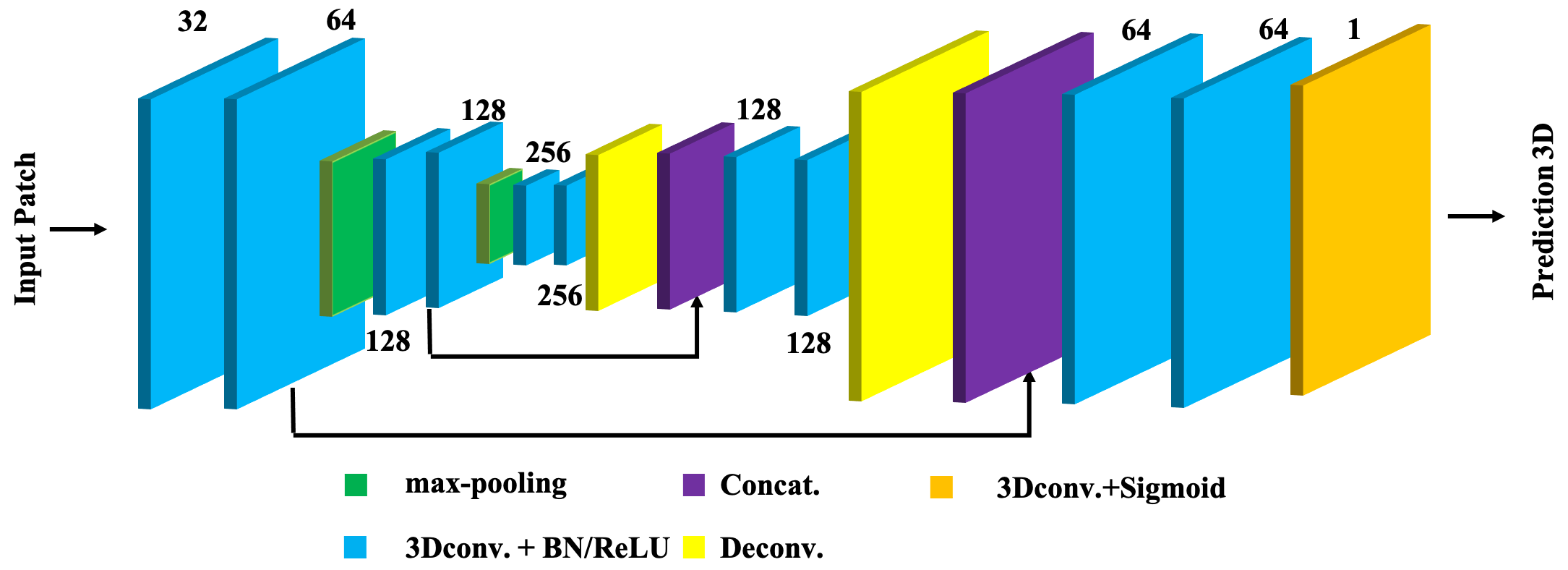}} 
\caption{Overview of the backbone UNet. The architecture is simplified for the patch-based binary segmentation in 3D US images.}
\label{backbone}
\end{figure}

We implemented our framework in Python 3.7 with \textit{TensorFlow} 1.10, using a standard PC with a TITAN 1080Ti GPU. We consider the Compact-UNet as the backbone architecture~\cite{MEDIA2020} in both branches, which has been proven to be successful in segmenting the instrument in 3D volumetric data, as shown in Fig.~\ref{backbone}. {We empirically reduce the number of scales and filter numbers of the UNet structure, which is to reduce the GPU memory cost and to fit the input patch size. In addition, this compact model could reduce overfitting with fewer trainable parameters ($\sim$4.5 million parameters). More details are shown in Table.~\ref{CNNPara}. As shown in the later experiment, this simplified CNN achieves comparable results to the standard one in our task.} It is worth to mention that although a common teacher-student framework has asymmetric design for knowledge distillation, our design considers a fair case where two networks are teaching each other for SSL.

\begin{table}[htbp]
\scriptsize

\centering
\caption{Architecture of the considered Compact-UNet. --[] denotes the skipping concatenation connection. The second column indicates the output of the current stage. $3^3$, 32, stride 1, BN/ReLU means 3D convolution with kernel size $3\times3\times3\times32$ with stride=1 with batch normalization and ReLU operations.}
\label{CNNPara}
\begin{tabular}{l|c|c}
\hline
&Feature size&Compact-UNet\\ \hline
input&$48^3$&-\\ 
convolution 1&$48^3$&$3^3$, 32, stride 1, BN/ReLU \\ 
convolution 2&$48^3$&$3^3$, 64, stride 1, BN/ReLU \\ 
pooling&$24^3$&$3^3$ max-pooling, stride 2\\
convolution 3&$24^3$&$3^3$, 128, stride 1, BN/ReLU \\ 
convolution 4&$24^3$&$3^3$, 128, stride 1, BN/ReLU \\ 
pooling&$12^3$&$3^3$ max-pooling, stride 2\\
convolution 5&$12^3$&$3^3$, 256, stride 1, BN/ReLU \\ 
convolution 6&$12^3$&$3^3$, 256, stride 1, BN/ReLU \\ 
deconvolution 1&$24^3$&$2^2$, stride 2,--[convolution 4]\\
convolution 7&$24^3$&$3^3$, 128, stride 1, BN/ReLU \\ 
convolution 8&$24^3$&$3^3$, 128, stride 1, BN/ReLU \\ 
deconvolution 2&$48^3$&$2^2$, stride 2,--[convolution 2]\\
convolution 9&$48^3$&$3^3$, 64, stride 1, BN/ReLU \\ 
convolution 10&$48^3$&$3^3$, 64, stride 1, BN/ReLU \\ 
Output&$48^3$&$1^3$, 1, stride 1, Sigmoid \\ \hline
Training setups & Parameter&Values\\ \hline
Adam optimizer & learning rate&1e-4\\
Weight parameter&$(\alpha, \beta,\gamma)$&(0.1, 0.002,0.1) for Gaussian ramp-up\\
Threshold parameter&$(\tau_1, \tau_2)$&(0.5, 0.7) for Gaussian ramp-up\\\hline
\end{tabular}
\end{table}

As for SSL training, the patches are generated by applying random translations based on the annotated instrument center points. Moreover, the data augmentations with rotation, mirror and intensity re-scales are applied. To adapt the UNet as a Bayesian network \cite{Bayesian} and generate uncertainty prediction, dropout layers with rate 0.5 were inserted prior to the convolutional layers. Gaussian random noise was also considered during uncertainty estimation. For the uncertainty estimation suggested by \cite{MICCAI2019SSL2}, $T=8$ was used to balance the efficiency and quality of the estimation. \emph{Ex-vivo} dataset training was terminated after the loss converged on the validation dataset or after 10,000 steps with a mini-batch size of 4 using the Adam optimizer (learning rate 1e-4)\cite{Adam}, which includes 2 labeled and 2 unlabeled patches. Meanwhile, the training on the \emph{in-vivo} dataset was terminated after the loss converged on the validation dataset or after 5,000 steps with a mini-batch size of 4 (learning rate 1e-4). Hyper-parameters $\alpha$, $\beta$ and $\gamma$ are empirically chosen as 0.1, 0.002 and 0.1 to balance different loss components. In addition, a ramp-up weighting coefficient strategy is considered for weight parameters to balance the components confidence during the training, which ables to ensure the objective loss is dominated by the supervised loss term. {Specifically, the ramp-up is set to be $\lambda\exp(-5(1-t/t_{max})^2)$, where $\lambda$ is parameter, $t$ is training steps and $t_{max}$ is set to be 4k~\cite{MICCAI2019SSL2}.} This avoids the learning processes get stuck in a case that no meaningful prediction of unlabeled data is obtained. In addition, we also consider Gaussian ramp-up paradigm~\cite{MICCAI2019SSL2},  to ramp up the uncertainty thresholds from $\frac{3}{4}\tau_1$ and $\frac{3}{4}\tau_2$ to $\tau_1$ and $\tau_2$, respectively (total step is 4k). The maximized thresholds values are obtained based on an uncertainty function with probability as 0.5 and 0.7 w.r.t. uncertainty estimation $U=-\sum_{c}p\text{log}(p)$. The $\tau_1$ employs 0.5 to have the maximized uncertainty in the both networks while $\tau_2$ employs a more tighter value to filter out more samples for a better estimation, which are empirically selected~\cite{DualStudent}. By doing so, our method would filter out less and less samples and enable the networks to gradually learn from the relatively certain to uncertain cases. To perform SSL, two individual networks are trained separately, which are then followed by the optimization of the discriminator of Eqn.~\ref{clss} in each iteration. The total training time for the two datasets were 14 and 7 hours, respectively. In all the dataset, the manual annotations are used as the ground truth for the evaluation.
\section{Results}\label{results}
Based on the DQN pre-selection, the instrument center can be localized with detection error of $3.8\pm1.8$ voxels and $2.4\pm1.0$ voxels for RF-ablation catheter dataset and TAVI guide-wire dataset, respectively. In contrast to our previous results in MICCAI, these values are obtained based on the resize input volume size as $96^3$ voxel, which leads to higher accuracy with a faster average detection time of around 0.2 seconds per volume. With the detected instrument center point, patches with the size of $48^3$ voxels are extracted around the point for semantic segmentation (i.e., 2 patches for each direction and $2^3$ patches in total). Performance comparison with the state-of-the-art methods and ablation studies are presented as follows (All the segmentation results are obtained based on the coarse localization). To evaluate the overall segmentation performance of the proposed method, we consider the Dice score (\textbf{DSC}) and $95\%$ Hausdorff Distance (\textbf{95HD}) as evaluation metrics~\cite{Metric}.
\subsection{Comparisons with existing methods}
\begin{figure}[htbp]

\centering{\includegraphics[width=7cm]{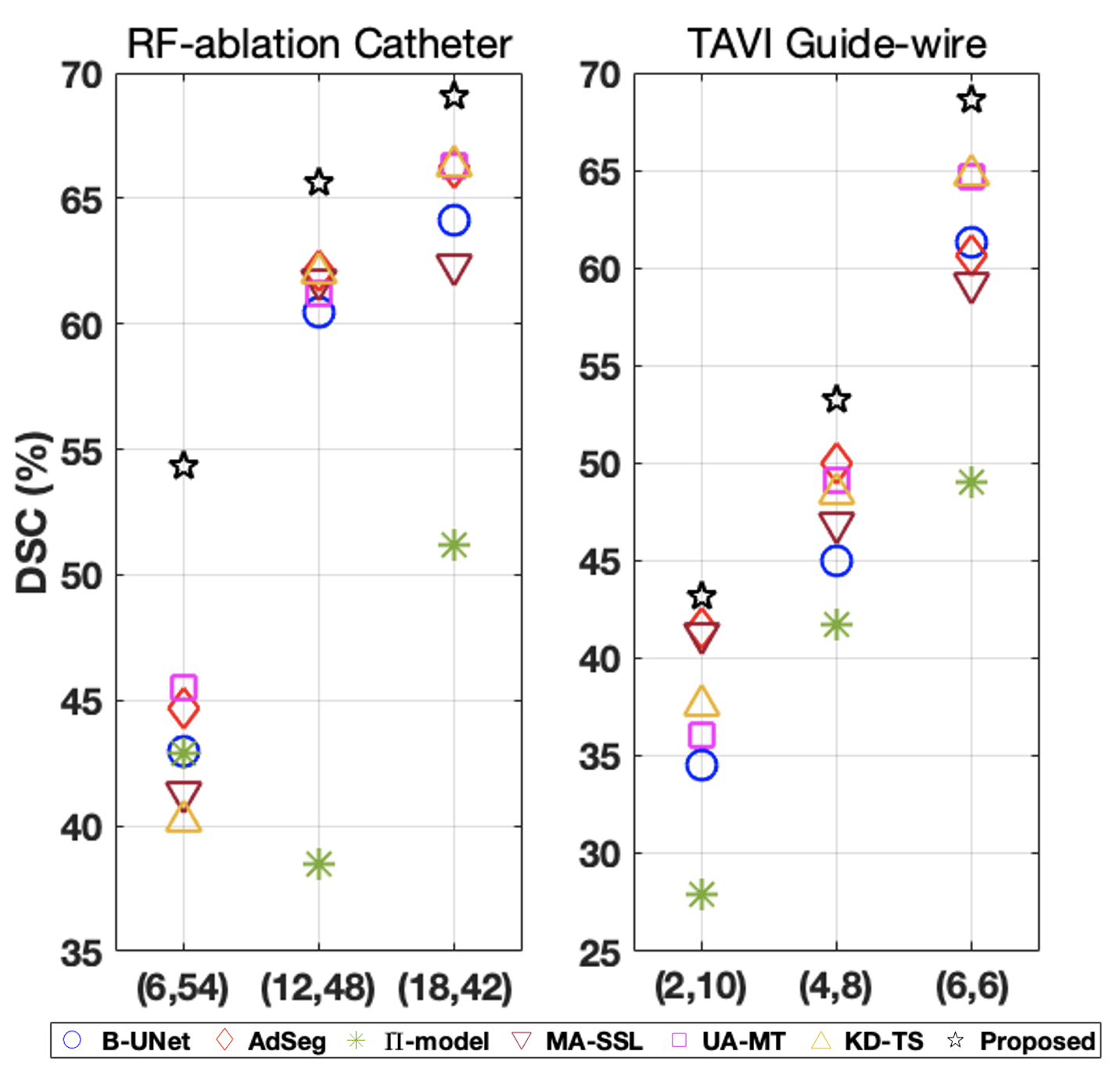}} 
\caption{Comparison with different methods for different (L,U) combination on two datasets. Different symbols represent different models, while the best results are also shown in Table~\ref{Compare}. Note the B-UNet considers no unlabeled images for training.}
\label{Unlabel_compares}
\end{figure}
\begin{figure}[htbp]
\centering{\includegraphics[width=7cm]{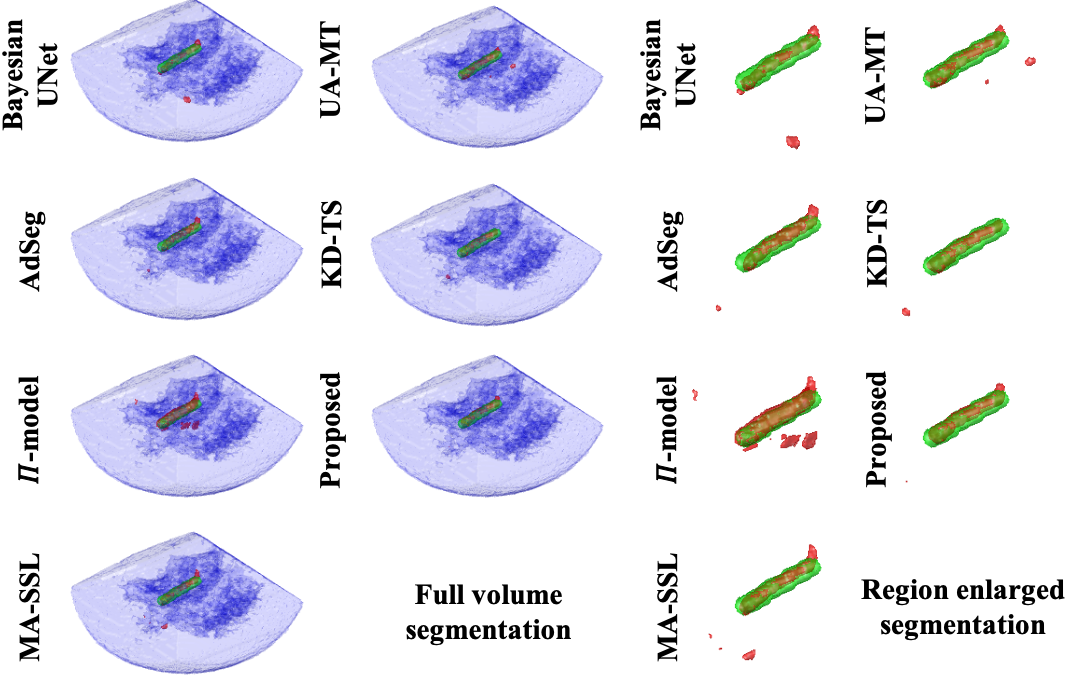}} 
\caption{Examples results via different methods for (L,U)=(18,42), which are corresponding to the Table~\ref{Compare}. Left: full volume, right: the enlarged region includes the catheter. Green: annotation, red: segmentation and blue: heart tissue. All the results are obtained based on the coarse detection.}
\label{compares}
\end{figure}

We compare the proposed method with the state-of-the-art SSL methods, including Bayesian UNet (B-UNet) \cite{MC}, $\Pi$-model \cite{BMVC2018} Adversarial-based segmentation (AdSeg)\cite{MICCAI2017Zhang}, multi-task attention-based SSL (MA-SSL)\cite{MASSL}, uncertainty-aware-based mean-teacher (UA-MT),\cite{MICCAI2019SSL2} and teacher-student-based (TS) knowledge distillation \cite{hinton2015distilling,zhou2020deep}. {Specifically, for methods of B-UNet and AdSeg, the backbone Compact-UNet is considered for fair comparison. As for B-UNet, Monte Carlo dropout is included to generate Bayesian estimation, which is same as our implementation in the proposed method. As for AdSeg, the adversarial classifier is implemented based on the Fig.~\ref{classifier}, which ensures the same adversarial classification procedure as our method. For MA-SSL, the method is implemented based on Compact-UNet with duplicated decoders for reconstruction and segmentation tasks. For the $\Pi$-model, the Dual-UNet is considered with EMA parameter updating, which is employed as the mean-teacher backbone. In addition,  random spatial transformation is applied based on the original implementation in 3D format~\cite{BMVC2018}. As for UA-MT, Dual-UNet with EMA updating is considered as the backbone, which employs the same uncertainty estimation of our method in Enq.~(\ref{LC}).  As for TS model, it trains the teacher part with a more complex model by increasing the filter number by a factor of two. The teacher is firstly trained on labeled images, which is then used to generate the soft-prediction of unlabeled images for the student model. The soften parameter of the unlabeled image is set to be 5 with loss weight of 0.5. Moreover, to ensure fair comparisons, the patch size, image augmentation, optimization steps are exactly the same as our method.} {Results are shown in Fig.~\ref{Unlabel_compares} and Table~\ref{Compare}, which depict that the proposed method outperforms the SOTA SSL approaches for the different (L,U) settings.} Examples of segmentation results of different SSL methods are shown in Fig.~\ref{compares}, where 18 annotated images were used for training. It can be observed that the proposed method provides fewer outliers than others because of better discriminative information exploration (all of them are obtained based on the DQN pre-selection). 

\begin{table}[htbp]
\scriptsize
\centering
\caption{Segmentation performance for different methods in DSC and 95HD, which are shown in mean$\pm$std. (L,U) means (Labeled, Unlabeled) images for SSL. All methods are based on DQN pre-selection results. The proposed method is noted as bold.}
\label{Compare}	
\begin{threeparttable}
\begin{tabular}{l|c|c|c}
\hline
\multirow{2}{*}{Method}& \multicolumn{1}{c|}{\emph{\# Images}}&\multicolumn{2}{c}{\emph{RF-ablation Catheter}}\\ \cline{2-4} 
& (L,U)& DSC \%& {95HD (voxels)}\\ \hline
B-UNet\cite{MC}&(18,0)&64.1$\pm$9.8${^{**}}$&{5.9$\pm$5.0}\\
AdSeg\cite{MICCAI2017Zhang}&(18,42)&66.2$\pm$8.7${^{**}}$&{10.3$\pm$11.1}\\
$\Pi$-model\cite{BMVC2018}&(18,42)&{51.2$\pm$12.5$^{**}$}&{$14.1\pm$8.2}\\
MA-SSL\cite{MASSL}&(18,42)&62.3$\pm$13.0${^{**}}$&{7.9$\pm$8.0}\\
UA-MT\cite{MICCAI2019SSL2}&(18,42)&66.3$\pm$9.2${^{**}}$&{4.3$\pm$3.2}\\
 {KD-TS \cite{hinton2015distilling}}&(18,42)& {66.3.$\pm$8.5}${^{**}}$&{4.6$\pm$4.8}\\
\textbf{Proposed}&(18,42)&\textbf{69.1$\pm$7.3}&\textbf{ {3.0$\pm$2.1}}\\\hline
Share-CNN\cite{IJCARS2019}&(60,0)&58.4$\pm$12.6${^{**}}$&{8.0$\pm$6.4}\\
Compact-UNet\cite{MEDIA2020}&(60,0)&66.8$\pm$7.3${^{*}}$&{4.0$\pm$3.1}\\
{Complex-UNet\cite{yang2018towards}}&{(60,0)}&{67.5$\pm$6.1${^{ns}}$}&{4.3$\pm$5.1}\\
Dual-UNet&(60,0)&{69.4$\pm$6.5}${^{ns}}$&{{3.6$\pm$3.0}}\\
Pyramid-UNet\cite{MEDIA2020}&(60,0)&\textbf{70.6$\pm$6.5}${^{ns}}$&\textbf{ {3.0$\pm$2.3}}\\\hline
\multirow{2}{*}{Method}&\multicolumn{1}{c|}{\emph{\# Images}}&\multicolumn{2}{c}{\emph{TAVI Guide-wire}}\\ \cline{2-4} 
& (L,U)& DSC \%& {95HD (voxels)}\\ \hline
B-UNet\cite{MC}&(6,0)&61.3$\pm$9.4${^{**}}$&{4.2$\pm$5.5}\\
AdSeg\cite{MICCAI2017Zhang}&(6,6)&60.6$\pm$7.7${^{*}}$&{5.3$\pm$5.2}\\
$\Pi$-model\cite{BMVC2018}&(6,6)&{49.0$\pm$9.3$^{**}$}&{5.1$\pm$1.2}\\
MA-SSL\cite{MASSL}&(6,6)&59.2$\pm$3.2${^{*}}$&{3.5$\pm$4.2}\\
UA-MT\cite{MICCAI2019SSL2}&(6,6)&64.7$\pm$7.3${^{**}}$& {1.8$\pm$0.6}\\
 {KD-TS \cite{hinton2015distilling}}&(6,6)&{64.7$\pm$8.1}${^{ns}}$&{2.2$\pm$1.2}\\
\textbf{Proposed}&(6,6)&\textbf{68.6$\pm$7.9}&\textbf{ {1.7$\pm$0.6}}\\\hline
Share-CNN\cite{IJCARS2019}&(12,0)&56.4$\pm$13.3${^{**}}$&{5.7$\pm$6.6}\\
Compact-UNet\cite{MEDIA2020}&(12,0)&63.2$\pm$6.6${^{*}}$&{1.9$\pm$0.5}\\
{Complex-UNet\cite{yang2018towards}}&{(12,0)}&{64.5$\pm$8.7${^{ns}}$}&{1.7$\pm$0.8}\\
Dual-UNet&(12,0)&{65.6$\pm$4.0}${^{ns}}$&\textbf{{1.5$\pm$0.2}}\\
Pyramid-UNet\cite{MEDIA2020}&(12,0)&\textbf{67.4$\pm$6.4}${^{ns}}$&\textbf{ {1.5$\pm$0.5}}\\\hline
\end{tabular}
\begin{tablenotes}
t-test between methods to ours: ns is p$>$0.05, * is {p$<$0.05}, ** is {p$<$0.01}.
\end{tablenotes}
\end{threeparttable}
\end{table}

From the figure, it can be seen that as the number of labeled images increases, the segmentation performances are improved w.r.t available supervised information. {By comparing our method to other SSL approaches, our method achieves the best performance. Specifically, for B-UNet, although it considers the Bayesian operation in the training to generalize the learning, this method cannot exploit the unlabeled images, which lead to the much worse performance. As for AdSeg, which considers adversarial learning to exploit the unlabeled images for semi-supervised learning, it achieves comparable results to other methods from literature. Nevertheless, due to its nature of exploiting the information at image level, the details of the predictions cannot be fully leveraged, which leads to worse results than ours. As for $\Pi$-model, it employs the spatial transformation to the mean-teacher architecture, which is similar to the UA-MT with same the backbone. Nevertheless, by comparing these two methods, the UA-MT achieves a better result, as it is focusing on the selected voxels for fine segmentation. These results indicate the voxel stability selection is important for our challenging task in the low contrast US imaging. Comparing UA-MT to KD-TS, although they utilize different approaches to smooth the prediction for knowledge transformation, their overall performances are comparable. And it is worth to mention that the soft information by temperature re-scale can enforce the KD-TS to better exploit the stable voxels, which is similar to the uncertainty estimation. As for MA-SSL, which exploits the SSL by an attention-based image reconstruction, it has worse performance than UA-MT, which also indicates the voxel selection is important for unsupervised signal exploration. As for our method, it achieves consistently the best performance for different (L, U) settings, since it exploits uncertainty and contextual information by the hybrid constraints, which improves the information usage of unlabeled images. Comparing results as shown in the ablation study section, the proposed hybrid constrains focus much more on aspects of the unlabeled images than the above state-of-the-art methods, which leads to better results in our task.} To further validate the performance differences among different SSL methods, we performed paired t-test with $\alpha=0.05$ on two datasets by DSC metric in the one-tailed test, which are summarized in Table~\ref{Compare} {for cases (18,42) and (6,6)}. From the table, the proposed method has larger statistical difference than other SSL methods on \emph{RF-ablation Catheter}, i.e., p-value$<$0.01. In contrast, the proposed method has a less statistical difference on \emph{TAVI Guide-wire}, especially for the KD-TS model, because of the limited testing dataset with only 4 images.

To further validate the our information exploration capacity for unlabeled images, we also compared the proposed method with the supervised learning methods. As shown in the bottom of Table~\ref{Compare}, the proposed method obtains better results than voxel-wise Share-CNN for catheter segmentation \cite{IJCARS2019}, which classifies voxels by CNN. Our method also outperforms the supervised learning method with backbone structure Compact-UNet, while it achieves similar performance to a more complex Pyramid-UNet {and a standard UNet (denoted as Complex-UNet)~\cite{yang2018towards}}. Meanwhile, by considering the fully supervised learning for Dual-UNet, the proposed SSL framework achieves comparable performances. Since proposed method is an SSL with multi-task learning, which exploits the information for different tasks to learn the discriminative knowledge in unlabeled images. It is worth to mention that the proposed method is statistically better than Share-CNN (p-value$<$0.01), while it has less difference with Compact-UNet (p-value$\sim$0.03). Comparing the standard UNet, Dual-UNet and Pyramid-UNet, there are no statistical differences from observation. These results show the proposed SSL method achieves the state-of-the-art performance with much less annotation effort than supervised learning methods, which follows the same coarse-to-fine framework. 

Although the Dual-UNet has a more complex architecture, which is around $9.2\times10^6$ parameters and $34.6\times10^{10}$ FLOPs for each patch at inference, it employs unlabeled images as the support and guidance for SSL training, which achieves comparable performances with fully supervised learning. Finally, from the experiments {(Python 3.7 with TensorFlow 1.10, using a standard PC with a TITAN 1080Ti GPU)}, the proposed two-stage scheme executes in around 1 second per volume (0.2-0.3 seconds for DQN pre-selection and 0.7 seconds for patch-based segmentation). A voxel-of-interest-based CNN method takes around 10 seconds~\cite{IJCARS2019}, while patch-of-interest coarse-to-fine segmentation spends around 1 second per volume \cite{MEDIA2020}. Therefore, our proposed method achieves the comparable efficiency to state-of-the-art instrument segmentation method in 3D US images.

\subsection{Ablation study of different loss components}
\begin{figure}[htbp]

\centering{\includegraphics[width=7cm]{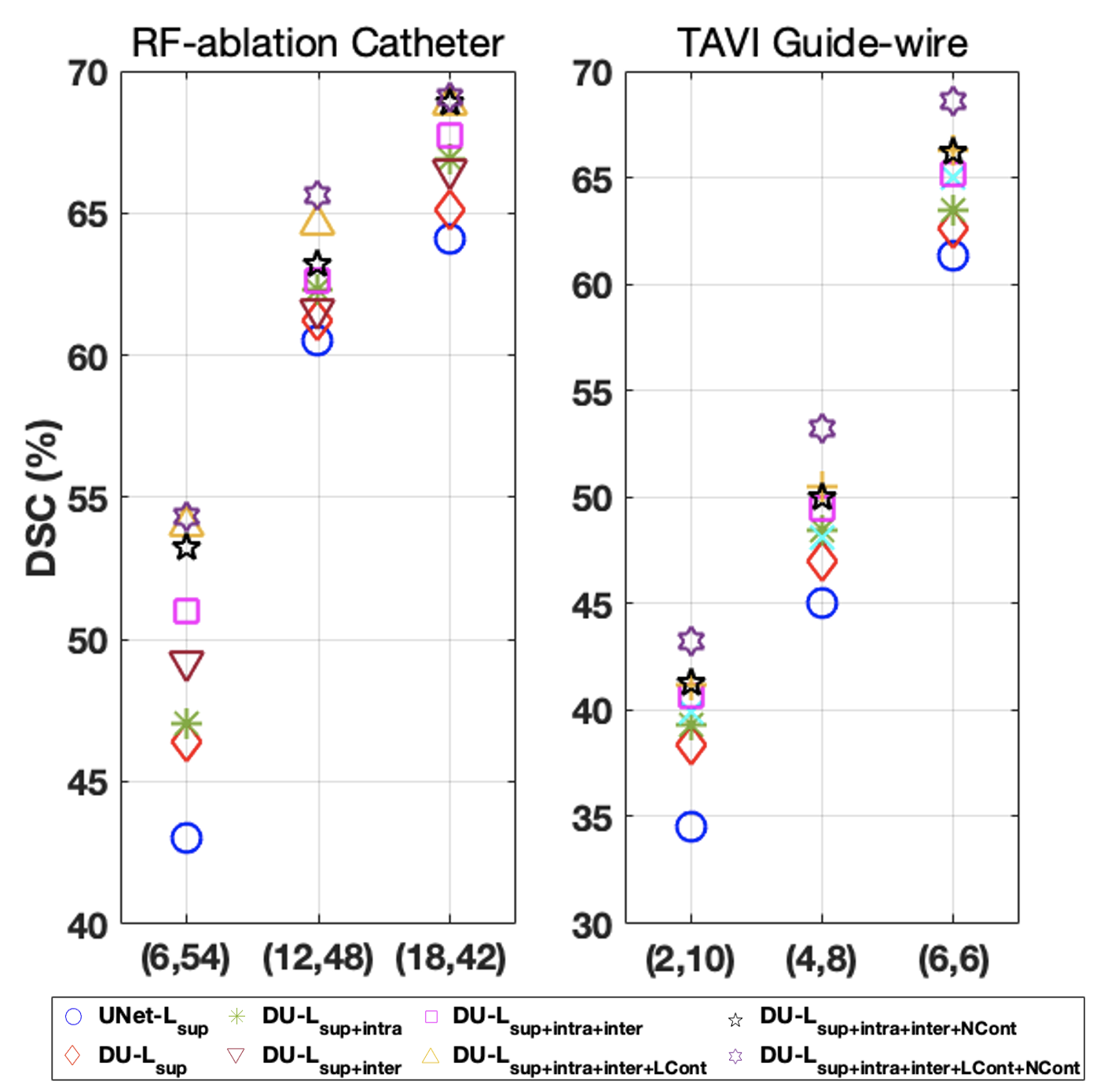}} 
\caption{Ablation study for different (L,U) combination on two datasets. Different symbols represent different models, which are corresponding to the methods in Table~\ref{ablation}. The best cases are also shown in the table.}
\label{ablation_compare}
\end{figure}

The ablation studies on different constraint components are summarized in {Fig.~\ref{ablation_compare}} where different numbers of labeled and unlabeled images are considered. More specifically, the UNet with Mont Carlo operation is denoted as the baseline and backbone structure for the proposed method, which is trained by $L_\text{sup}$. For the proposed SSL, constraint components are added one by one to validate their effectiveness. {The numerical results of the best performances of this ablation study are shown in Table~\ref{ablation}.}

Several conclusions can be drawn from the figure and table. (1) The simple backbone UNet with supervised loss can learn more discriminating information with the number of available annotations increasing, which however obtains worse performance than the Dual-UNet. This is because randomly initialized parameters and dropout operations in Dual-Unet avoid the learning bias with higher network diversity, which results in more stable predictions. {(2) Compared to the case with only supervised loss, adding voxel-level constraints, i.e., $L_\text{intra}$ and $L_\text{inter}$, allows to select the stable voxels from uncertainty estimations, which therefore exploits the discriminating information from unlabeled images’ prediction. More specifically, $L_\text{intra}$ constraint focuses on prediction uncertainty within the network while the $L_\text{inter}$ exploit the uncertainties of the predictions between two individual networks. The results indicate both constraints improve the performance and are complementary to each other. (3) The contextual-level constraint, including label-wise, and network-wise constraints, also contribute to further performance improvement. Specifically, the label-wise constraint exploits the contextual similarity between labeled and unlabeled images’ predictions, while network-wise constraint focuses on prediction similarity between different networks of Dual-UNet. (4) The proposed hybrid loss has more significant performance improvement when the amount of labeled images are small, which indicates our proposal is able to exploit the discriminating information from unlabeled images.} It is worth to mention that in our implementation, both networks are considering a same backbone Compact-UNet due to limited GPU memory and consideration of efficiency. In realistic, they can have different network structures with more than two branches ~\cite{DualStudent}. However we failed to observe performance improvement. 

\begin{table}[htbp] 
\scriptsize
\centering
\caption{Segmentation performance for loss components in DSC and 95HD, which are shown in mean$\pm$std. (L,U) means (Labeled, Unlabeled) images for SSL training. DU means Dual-UNet. The best performances are bolded.}
\label{ablation}	
\begin{tabular}{l|c|c|c}
\hline
\multirow{2}{*}{Method}& \multicolumn{1}{c|}{\emph{\# Images}}& \multicolumn{2}{c}{\emph{RF-ablation Catheter}}\\ \cline{2-4} 
& (L,U)& DSC \%&{95HD (voxels)}\\ \hline
UNet-$L_\text{sup}$&(18,0)&64.1$\pm$9.8&{5.9$\pm$5.0}\\
DU-$L_\text{sup}$&(18,0)&65.1$\pm$8.9&{5.5$\pm$3.5}\\
DU-$L_\text{sup+intra}$&(18,42)&66.9$\pm$7.7&{4.9$\pm$5.0}\\
DU-$L_\text{sup+inter}$&(18,42)& {66.5$\pm$9.3}&{5.2$\pm$6.6}\\
DU-$L_\text{sup+intra+inter}$&(18,42)&67.7$\pm$8.5&{3.5$\pm$2.2}\\
DU-$L_\text{sup+intra+inter+LCont}$&(18,42)&68.8$\pm$7.2&{3.3$\pm$2.2}\\
DU-$L_\text{sup+intra+inter+NCont}$&(18,42)& {68.9$\pm$7.5}&{{4.1$\pm$3.3}}\\
DU-$L_\text{sup+intra+inter+LCont+NCont}$&(18,42)&\textbf{69.1$\pm$7.3}&\textbf{ {3.0$\pm$2.1}}\\\hline
\multirow{2}{*}{Method}&\multicolumn{1}{c|}{\emph{\# Images}}&\multicolumn{2}{c}{\emph{TAVI Guide-wire}}\\ \cline{2-4} 
& (L,U)& DSC \%&{95HD (voxels)}\\ \hline
UNet-$L_\text{sup}$&(6,0)&61.3$\pm$9.4&{4.2$\pm$5.5}\\
DU-$L_\text{sup}$&(6,0)&62.6$\pm$6.6&{2.2$\pm$1.1}\\
DU-$L_\text{sup+intra}$&(6,6)&63.5$\pm$10.5&{2.9$\pm$1.3}\\
DU-$L_\text{sup+inter}$&(6,6)& {65.0$\pm$9.4}&{2.4$\pm$0.8}\\
DU-$L_\text{sup+intra+inter}$&(6,6)&65.2$\pm$8.0&{1.9$\pm$1.0}\\
DU-$L_\text{sup+intra+inter+LCont}$&(6,6)&66.3$\pm$5.2&{1.8$\pm$0.4}\\
DU-$L_\text{sup+intra+inter+NCont}$&(6,6)& {66.2$\pm$9.0}&\textbf{ {1.6$\pm$0.5}}\\
DU-$L_\text{sup+intra+inter+LCont+NCont}$&(6,6)&\textbf{68.6$\pm$7.9}&{1.7$\pm$0.6}\\\hline
\end{tabular}
\end{table}

\begin{figure}[htbp]

\centering{\includegraphics[width=5cm]{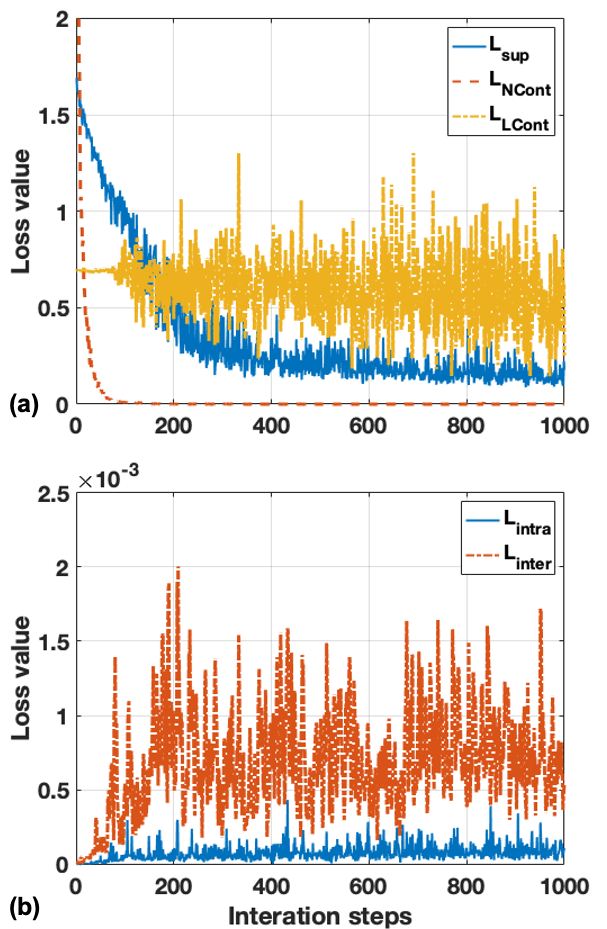}} 
\caption{Training loss curves of different loss components. (a) Curves of $L_\text{sup}$, $L_\text{NCont}$ and $L_\text{LCont}$. (b) Curves of $L_\text{intra}$, $L_\text{inter}$.}
\label{losscurve}
\end{figure}

It can be observed that as the number of annotated images increases, the variance of the segmentation performance is decreasing; this is because more confident guidance is obtained from available annotations. In the following ablation studies, we have chosen the cases with the most annotated volumes for both datasets, i.e., (18,42) and (6,6) combinations for labeled and unlabeled training images. {Based on case of (18,42) for \emph{ex-vivo} dataset, training loss curves are depicted in Fig.~\ref{losscurve}. As can be observed in (a), $L_\text{sup}$ and $L_\text{NCont}$ are consistently reduced as iteration proceeding. In contrast, $L_\text{LCont}$ is fluctuating around 0.6, as it is used to force the labeled and unlabeled predictions to be the same. It is worthwhile to notice the value of $L_\text{NCont}$ converges to a small value rather than 0. From figure (b), $L_\text{inter}$ has larger loss values than $L_\text{intra}$, as it focuses on the voxels from two different networks, and it is more difficult to minimize the prediction differences. The values of these two losses increases at the beginning of the training, which is due to more voxels are selected as the iteration proceeding.}  

\subsection{Ablation study of the parameters in loss components}
\begin{figure*}[htbp]
\centering{\includegraphics[width=18cm]{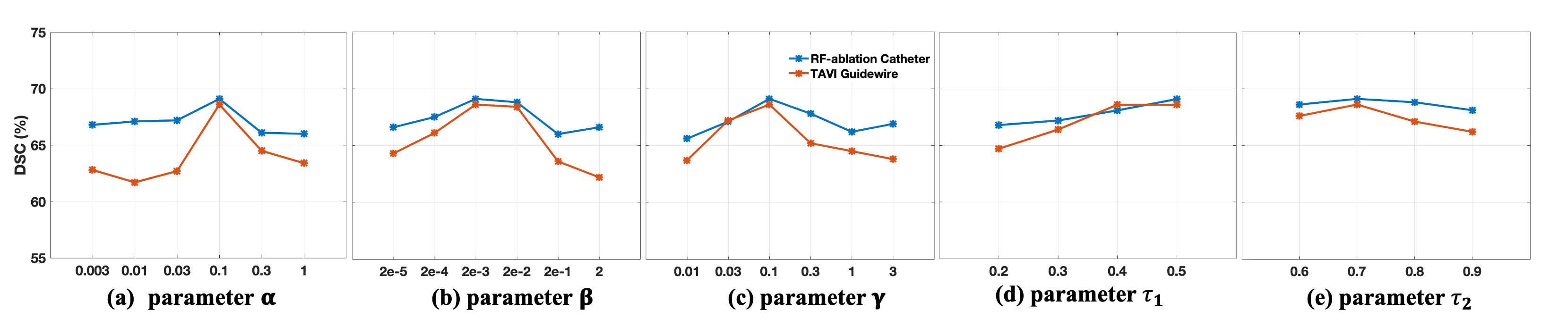}} 
\caption{Segmentation performances based on different hyper parameters in Eqn.(3), (4) and (8). Note that values for $\tau_1$ and $\tau_2$ are probability.}
\label{weights}
\end{figure*}

Experiments were performed to investigate whether the performance is sensitive to the weights $\alpha$, $\beta$ and $\gamma$ of hybrid loss. Three weights were tuned separately, which means for each experiment setting, one parameter was changed while the remaining two were fixed. The results in Fig.~\ref{weights} indicate the performance is sensitive to the parameters $\alpha$ and $\gamma$ while less sensitive to the parameter $\beta$ {given the range of 2e-3 to 2e-2. Nevertheless, it is more sensitive to the parameter $\beta$ on the TAVI Guide-wire dataset.} Similarly, threshold parameters $\tau_1$ and $\tau_2$ are also validated, which are less sensitive than the above three values. Nevertheless, lower $\tau_1$ and higher $\tau_2$ would lead to worse performances.

\begin{table}[tb] 
\centering
\scriptsize
\caption{Segmentation performance for different similarity metrics in $L_\text{NCont}$ in DSC and 95HD, which are shown in mean$\pm$std. The best performances are bolded.}
\label{space}	
\begin{tabular}{l|c|c}
\hline
\multirow{2}{*}{Method}& \multicolumn{2}{c}{\emph{RF-ablation Catheter}}\\ \cline{2-3} 
&DSC \%&95HD (voxels)\\ \hline
SRC&68.4$\pm$7.8&6.1$\pm$8.0\\ 
Cosine Distance&65.3$\pm$11.2&3.9$\pm$2.4\\ 
L2 Distance&\textbf{69.1$\pm$7.3}&\textbf{3.0$\pm$2.1}\\\hline 
\multirow{2}{*}{Method}& \multicolumn{2}{c}{\emph{TAVI Guide-wire}}\\ \cline{2-3} 
&DSC \%&95HD (voxels)\\ \hline
SRC&62.3$\pm$10.1&1.6$\pm$0.4\\ 
Cosine Distance&64.4$\pm$9.2&\textbf{1.4$\pm$0.3}\\ 
L2 Distance&\textbf{68.6$\pm$7.9}&{1.7$\pm$0.6}\\ \hline
\end{tabular}
\end{table}

\subsection{Ablation study of similarity measure in $L_{NCont}$}
{The similarity of feature vectors should be measured by a Riemannian distance with a proper Riemann metric \cite{arvanitidis2017latent}. Unfortunately, the Riemannian distance is difficult to measure in practice. Alternatively, we experientially compared three similarity measurements: Norm-2 distance, Cosine similarity \cite{chen2020simple} and Sample Relation Consistency (SRC) \cite{liu2020semi}. The results summarized in Table~\ref{space} show the Norm-2 distance has the best performance, while Cosine similarity provides the worse performance, since} {it only considers the direction similarity while ignoring the length difference. }

\subsection{Ablation study of patch size of Dual-UNet}
{To investigate the influence of the patch size, the input patch size of $32^3$, $48^3$ and $64^3$ were examined. The results are shown in Table~\ref{patch}. From the results, patches with $32^3$ voxels obtained a little bit worse performance than $48^3$ voxels, which however requires around 3 seconds due to more patches are required for a fixed volume size after DQN pre-selection ($64^3$ voxels). In contrast, patches with $64^3$ voxels have similar time to $48^3$ voxels (0.7 seconds) but obtained much worse performance with higher GPU memory usage (we set batch=1 for this case). Although a larger contextual information can be captured, it is more easily to be overfitted compared to a smaller patch size. As a result, the optimal patch size is $48^3$.}

\begin{table}[htbp]
\scriptsize
\centering
\caption{Ablation studies of different patch size for Dual-UNet. Performance are evaluated by DSC and 95HD, which are shown in mean$\pm$std. The best performances are bolded.}
\label{patch}
\begin{tabular}{l|c|c}
\hline
\multirow{2}{*}{Patch Size}& \multicolumn{2}{c}{\emph{RF-ablation Catheter}}\\ \cline{2-3} 
&DSC \%&95HD (voxels)\\ \hline
$32^3$ voxels&68.5$\pm$7.9&3.3$\pm$1.9\\ 
$48^3$ voxels&\textbf{69.1$\pm$7.3}&\textbf{3.0$\pm$2.1}\\ 
$64^3$ voxels&66.3$\pm$9.6&3.7$\pm$2.4\\\hline 
\multirow{2}{*}{Patch Size}& \multicolumn{2}{c}{\emph{TAVI Guide-wire}}\\ \cline{2-3} 
&DSC \%&95HD (voxels)\\ \hline
$32^3$ voxels&66.7$\pm$9.5&\textbf{1.6$\pm$0.3}\\ 
$48^3$ voxels&\textbf{68.6$\pm$7.9}&1.7$\pm$0.6\\ 
$64^3$ voxels&62.3$\pm$10.0&1.9$\pm$0.9\\ \hline
\end{tabular}
\end{table}

\subsection{Ablation study of pre-selection}
\begin{figure}[htbp]
\centering{\includegraphics[width=6cm]{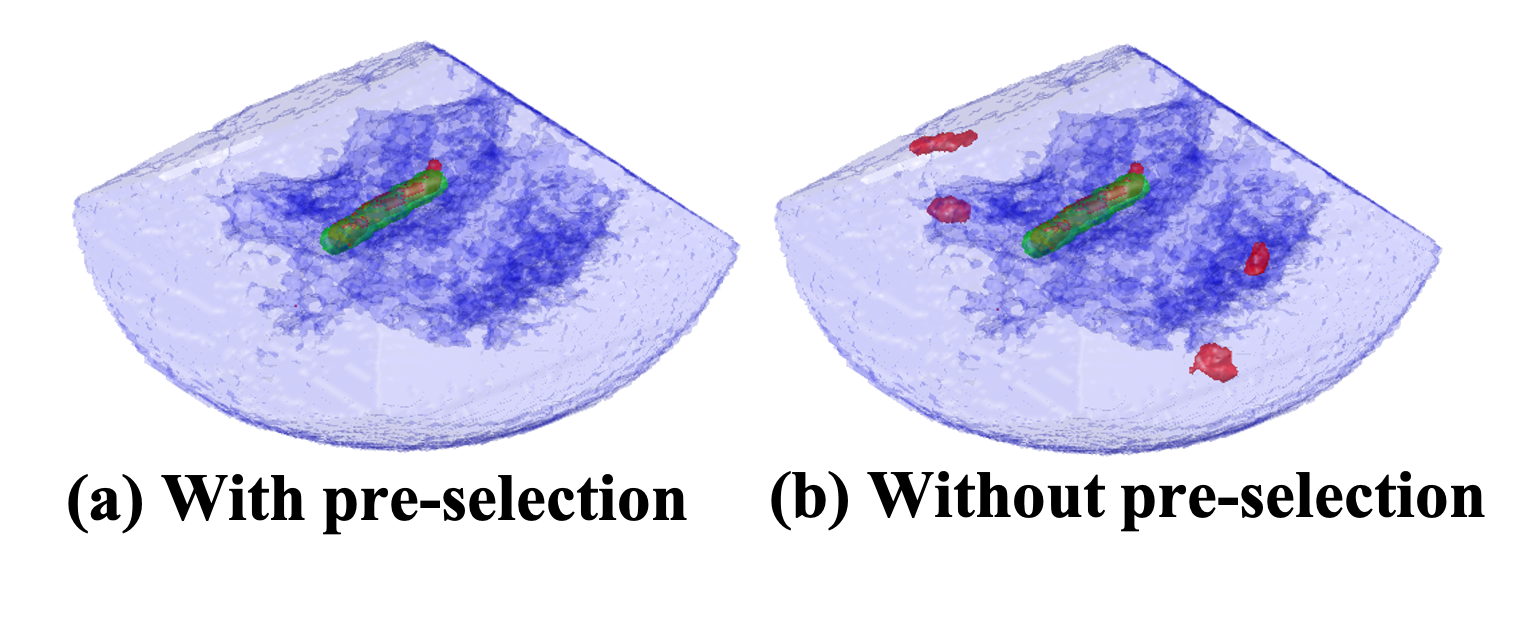}} 
\caption{Example volumes of the segmentation results with/-out pre-selection. Green: annotation, red: segmentation result.}
\label{DNQwith}
\end{figure}

{Experiment results with and without pre-selection were summarized in Table~\ref{sample}. As can be observed, the pre-selection improves the overall segmentation performance. Example images with and without  pre-selection are shown in Fig.~\ref{DNQwith}, which demonstrates the coarse pre-selection can omit the outliers outside the instrument region. In addition, the coarse pre-selection would drastically reduce the overall computational time for the instrument segmentation.}

\begin{table}[htbp]
\scriptsize
\centering
\caption{Ablation studies of coarse pre-selection. Segmentation performance are evaluated by DSC and 95HD, which are shown in mean$\pm$std. The best performances are bolded.}
\label{sample}
\begin{tabular}{l|c|c}
\hline
\multirow{2}{*}{Patch Size}& \multicolumn{2}{c}{\emph{RF-ablation Catheter}}\\ \cline{2-3} 
&DSC \%&95HD (voxels)\\ \hline
w/o selection&44.9$\pm$21.3&50.0$\pm$22.2\\ 
w selection&\textbf{69.1$\pm$7.3}&\textbf{3.0$\pm$2.1}\\ \hline
\multirow{2}{*}{Patch Size}& \multicolumn{2}{c}{\emph{TAVI Guide-wire}}\\ \cline{2-3} 
&DSC \%& 95HD (voxels)\\ \hline
w/o selection&57.8$\pm$13.9&32.3$\pm$22.3\\ 
w selection&\textbf{68.6$\pm$7.9}&\textbf{1.7$\pm$0.6}\\ \hline
\end{tabular}
\end{table}

\subsection{Generalization against different recording settings}
{To further validate the generalization of the proposed method,} {both DQN and SSL segmentation, the trained model from \emph{ex-vivo} RF-ablation catheter dataset (18 labeled images) is directly applied on the \emph{in-vivo} RF-ablation catheter dataset. The proposed coarse pre-selection successfully detects the catheter with the accuracy of $6.7\pm2.4$ voxel. Although it is a bit worse than the result on the \emph{ex-vivo} RF-ablation catheter dataset, it still can localize the catheter with 100\% successful rate, which shows the generalization of the DQN method. Based on the pre-selected regions, the Dual-UNet segmentation networks are applied to segment the catheter; the results are summarized in Table~\ref{livepig}. As can be observed, although the performances of the proposed method are degraded a bit, the overall performance is still reasonable. The proposed method produces better results than other state-of-the-art methods.}

\begin{table}[htbp]
\scriptsize
\centering
\caption{Segmentation performance for different methods on \emph{in-vivo} RF-ablation catheter dataset, which are evaluated by DSC and 95HD in mean$\pm$std. The proposed method is noted as bold.}
\label{livepig}
\begin{tabular}{l|c|c}
\hline
\multirow{2}{*}{Patch Size}& \multicolumn{2}{c}{\emph{in-vivo} RF-ablation catheter}\\ \cline{2-3} 
&DSC \%& 95HD (voxels)\\ \hline
B-UNet\cite{MC}&30.3$\pm$20.9&10.2$\pm$10.4\\ 
AdSeg\cite{MICCAI2017Zhang}&58.9$\pm$5.4&7.4$\pm$6.5\\ 
$\Pi$-model\cite{BMVC2018}&{45.6$\pm$9.2}&{13.5$\pm$7.1}\\ 
MA-SSL\cite{MASSL}&47.3$\pm$14.9&8.5$\pm$3.5\\ 
UA-MT\cite{MICCAI2019SSL2}&52.4$\pm$5.1&11.4$\pm$4.8\\ 
KD-TS \cite{hinton2015distilling}&52.8$\pm$7.4&10.3$\pm$4.0\\ 
\textbf{Proposed}&\textbf{63.8$\pm$10.1}&\textbf{6.2$\pm$5.1}\\\hline
\end{tabular}
\end{table}

\section{Limitation and discussions}\label{discussions}
Despite the above promising performances, there are still some limitations to our method. (1) The Monto Carlo method in the Bayesian network introduces random noise during the training, of which the uncertainty requires more training iterations to converge and stabilize. (2) {It is worth to mention that, although the statistical analysis shows the difference between different methods, the number of testing samples is limited, which are less than 30 images. A larger testing dataset is required for further validation in future.} So that a more complex and larger size dataset should be considered for further validation. (3) As stated in Dual-Student \cite{DualStudent}, the two individual networks can have different complexity and can even have more than three branches to learn the knowledge. However, due to the size of 3D UNets and computation complexity, it is difficult to achieve these forms on a GPU with limited memory. (4) As can be observed from the results of generalization analysis, the proposed method still has a performance degradation when applied to unseen datasets under different recording settings. {A recent study~\cite{zhuang2019self} has shown that the pre-trained self-supervised feature learning could improve the performance generalization for the segmentation, which can be considered as a research direction for improving the generalization and performance in future work. Nevertheless, it is worth to mention that unlike the application of brain tumor segmentations~\cite{zhuang2019self}, our task with the instrument in cardiac intervention has large variation of background tissue and arbitrary pose direction of the instrument (also due to data augmentations), which makes it challenging to learn the order and rotation information. (5) Finally, artifacts and speckle noises are commonly existing in US imaging, which would hamper the segmentation performance. Because of the difficulty of getting the in-vivo data, only limited \emph{in-vivo} data is used in our experiments. This is not enough for thorough validation of the proposed method for \emph{in-vivo} usage, as the noises are commonly existing in the clinical practice. Further \emph{in-vivo} data collection and validation will be performed in future to fully validate the effectiveness and robustness of the proposed method.}
\section{Conclusions}\label{conclusions}
{3D Ultrasound-guided therapy has been widely used, but it is difficult for a sonographer to localize the instruments in 3D US, because of the complex manual handling of instruments. Therefore, automated detection of medical instruments in 3D US is required to reduce the operation effort and thereby increase the efficiency. Nevertheless, it is expensive to train an automated deep learning method, which requires a large number of training images with careful annotations. In this paper,} we have proposed a SSL learning framework for instrument segmentation in US-guided cardiac interventions. The SSL method avoids intensive annotation effort while enabling the use of unlabeled images, which are guided by the advanced Dual-UNet with hybrid loss functions. The extensive comparison shows that the proposed method outperforms the state-of-the-art methods, while it achieves a comparable performance to the supervised learning approach with fewer annotations. 
\section*{{Appendix}}
\subsection{Gradients of $L_\text{semi}$ components}
{The proposed train method includes novel semi-supervised learning constraints. In terms of gradient descent-based optimization, their gradients w.r.t. prediction values are shown in the following (the supervised terms have been shown in Sudre \emph{et al.}~\cite{sudre2017generalised}). Formula $\mathcal{I}(\hat{U}<\tau_1)$ in Enq.~(\ref{LC}) is a binary mask for the prediction patch to select voxels. Therefore, this constraint, for each selected voxels, is equivalent to a L2 loss with a pre-calculated weight $\omega$ for prediction $\hat{y}_i$ and its averaged Bayesian estimation $\hat{p}_i$:}
\begin{equation}\label{voxels_1}
L_\text{intra}=\frac{\sum(\mathcal{I}(\hat{U}<\tau_1)\odot||\hat{y}-\hat{P}||)}{\sum\mathcal{I}(\hat{U}<\tau_1)}=\omega\times\sum(\hat{y}_i-\hat{p}_i)^2.
\end{equation}
{Therefore, its gradient w.r.t. to $\hat{y}_i$ is}
\begin{equation}\label{gradient1}
\frac{\partial{L_\text{intra}}}{\partial{\hat{y}_i}}=(\hat{y}_i-\hat{p}_i)\times\omega\times2
\end{equation}
{The formula is differentiable at the definition range. Similarly, $L_\text{inter}$ in Eqn.~(\ref{LS}) has similar definition for stable voxels to Eqn.~(\ref{voxels_1}) with different weight parameters, which therefore has similar gradient w.r.t. $\hat{y}_i$ in the Eqn.~(\ref{gradient1}). As for $L_{\text{LCont}}$, which employs commonly used binary cross-entropy, it is differentiable and its gradient w.r.t. $\hat{Cls}$ is defined as}
\begin{equation}\label{voxels_2}
\frac{\partial{L_\text{LCont}}}{\partial{\hat{Cls}}}=\frac{Cls}{\hat{Cls}}-\frac{1-Cls}{1-\hat{Cls}}.
\end{equation}

{Finally, for the constraint $L_{\text{NCont}}$, it measures the distance between two vectors of input ground truth and prediction, which are encoded by a contextual encoder. As for a ground truth vector $V(v_1,v_2,...,v_i,v_n)$ and prediction $\hat{V}(\hat{v}_i,\hat{v}_2,...,\hat{v}_i,...,\hat{v}_n)$ with length $n$, the Eqn.~(\ref{context}) can be re-formulated as} 
\begin{equation}\label{context_1}
\begin{split}
&L_{\text{NCont}}=||\hat{Y}_1^l-Y||+||\hat{Y}_2^l-Y||+||\hat{Y}_1^u-\hat{Y}_2^u||\\
&=\sum\left((\hat{v}_{1i}^l-v_i)^2+(\hat{v}_{2i}^l-v_i)^2+(\hat{v}_{1i}^u-\hat{v}_{2i}^u)^2\right),
\end{split}
\end{equation}
{where we avoid the square root of norm-2 for simplicity. Based on the above, its gradient w.r.t. to $\hat{v}^l_{ji}$ ($j$ represents for network 1 or 2 with labeled image) is defined as}
\begin{equation}\label{contex_2}
\frac{\partial{L_\text{NCont}}}{\partial{\hat{v}^l_{ji}}}=2\times(\hat{v}_{ji}^l-v_i).
\end{equation}
{Similarly, for unlabeled image, the gradient of Eqn.~(\ref{context_1}) w.r.t. $\hat{v}^u_{1i}$ is defined as}
\begin{equation}\label{contex_3}
\frac{\partial{L_\text{NCont}}}{\partial{\hat{v}^u_{1i}}}=2\times(\hat{v}_{1i}^u-\hat{v}_{2i}^u),
\end{equation}
{which is also validated for $\hat{v}^u_{2i}$. Based on the all above derivatives and chain rule, the overall joint training can be achieved by gradient descent methods.}
\bibliographystyle{IEEEtran}
\bibliography{draft}
\end{document}